\pdfoutput=1
\documentclass[11pt]{article}

\usepackage[]{old_acl}
\usepackage{url}
\usepackage{times}
\usepackage{latexsym}
\usepackage{comment}
\usepackage{enumitem}
\usepackage[linesnumbered,ruled,vlined]{algorithm2e}
\SetAlgorithmName{Alg.}{Alg.}{List of Algorithms}

\usepackage{amsmath}
\usepackage{wrapfig}
\usepackage[T1]{fontenc}
\usepackage[utf8]{inputenc}

\usepackage{microtype}
\usepackage{subcaption}
\usepackage{inconsolata}

\usepackage{graphicx}
\usepackage{booktabs}
\usepackage{multirow}

\title{\textsc{Infogent}: An Agent-Based Framework for Web Information Aggregation}

\author{Revanth Gangi Reddy\thanks{Equal Contribution.}\hspace{0.5em}Sagnik Mukherjee\footnotemark[1]\hspace{0.5em}Jeonghwan Kim\footnotemark[1]\hspace{0.5em}Zhenhailong Wang\footnotemark[1]\\\textbf{Dilek Hakkani-Tur}\hspace{1em}\textbf{Heng Ji}\\
University of Illinois at Urbana-Champaign \\
  \texttt{\{revanth3,sagnikm3,jk100,wangz3,dilek,hengji\}@illinois.edu}}

\begin{document}

\maketitle

\begin{abstract}
Despite seemingly performant web agents on the task-completion benchmarks, most existing methods evaluate the agents based on a presupposition: the web navigation task consists of linear sequence of actions with an end state that marks task completion. In contrast, our work focuses on web navigation for \textit{information aggregation}, wherein the agent must explore different websites to gather information for a complex query. 
We consider web information aggregation from two different perspectives: (i) \textit{Direct API-driven Access} relies on a text-only view of the Web, leveraging external tools such as Google Search API to navigate the web and a scraper to extract website contents. (ii) \textit{Interactive Visual Access} uses screenshots of the webpages and requires interaction with the browser to navigate and access information.
Motivated by these diverse information access settings, we introduce \textsc{Infogent}\footnote{Code will be available at \url{https://github.com/gangiswag/infogent}.}, a novel modular framework for web information aggregation involving three distinct components: Navigator, Extractor and Aggregator. Experiments on different information access settings demonstrate \textsc{Infogent} beats an existing SOTA multi-agent search framework by 7\% under Direct API-Driven Access on FRAMES, and improves over an existing information-seeking web agent by 4.3\% under Interactive Visual Access on AssistantBench.
\end{abstract}

\section{Introduction}
\label{sec:intro}
% \heng{need to summarize clearly what kind of requests can benefit from BFS? "Why" kind of queries? queries require information from multiple domains? indefinite scope of answers?}

\begin{figure}[t]
    \centering
    \includegraphics[width=0.9\linewidth]{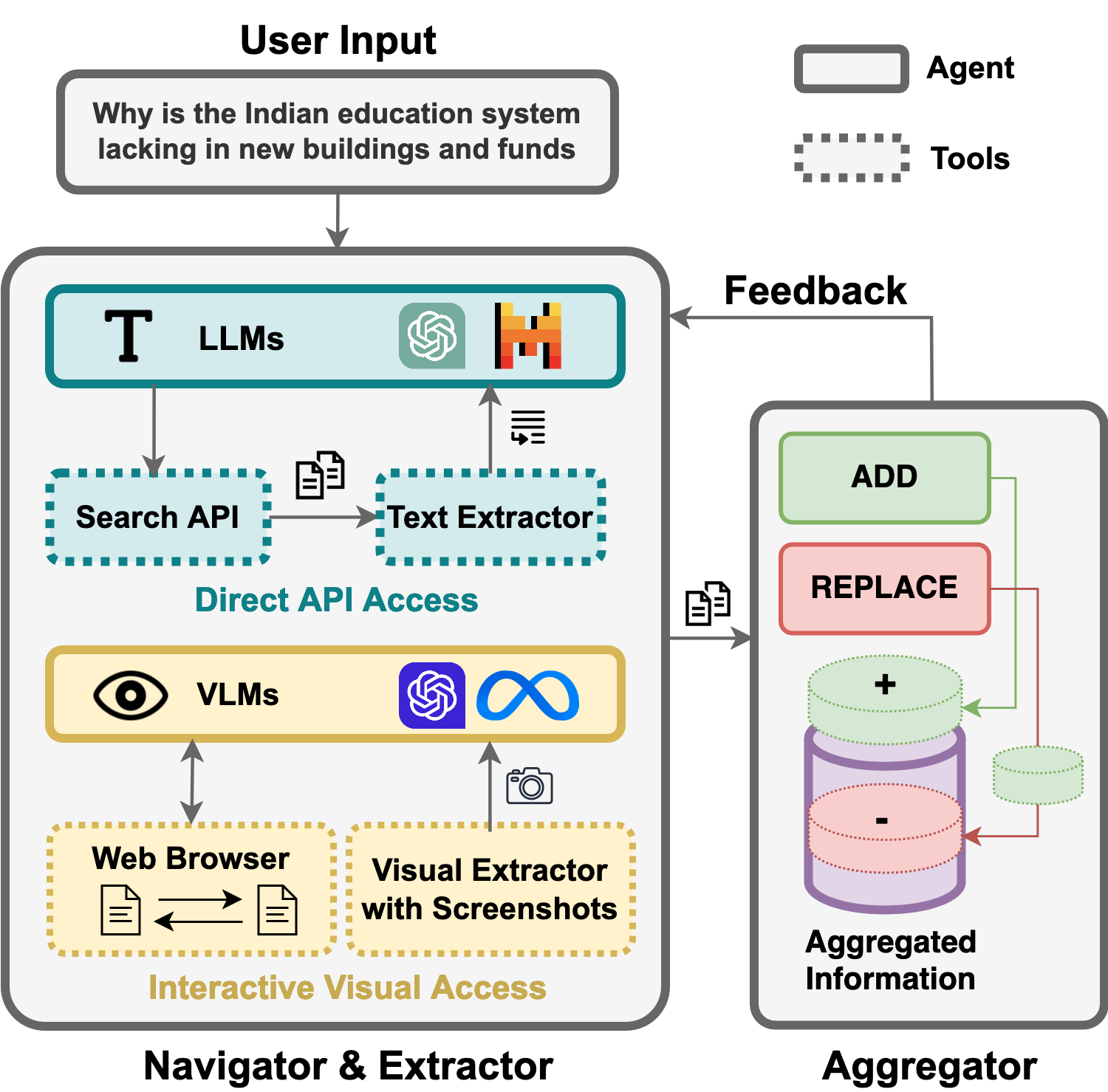}
    \vspace{-0.3em}
    \caption{Overview of \textsc{Infogent} under the \textit{Direct API Access} and \textit{Interactive Visual Access} settings: The Navigator uses a tool-based LLM and a browser-controlling VLM as the web agent respectively, with the Aggregator's textual feedback guiding further navigation.}
    \label{fig:infogent_overview}
    \vspace{-1em}
\end{figure}

Despite the well-documented success of autonomous web agents~\cite{nakano2021webgpt, yang2023autogpt,zhou2023webarena, deng2024mind2web}, the proposed tasks usually perform goal-oriented web-based tasks involving navigating within a website, interacting with elements like buttons and executing complex workflows. (e.g., booking a flight or scheduling a meeting). However, a critical aspect of web-based tasks, \textit{information aggregation}  has received relatively less attention. Tasks involving gathering and presenting relevant data from diverse web sources are central to many real-world applications. For instance, humans often visit multiple websites, using search engines to find relevant content, and browsing articles, reviews, or forums. 

Existing web navigation benchmarks and methods~\cite{zhou2023webarena, deng2024mind2web, lu2024weblinx, zheng2024seeact, koh2024visualwebarena} primarily focus on linear, goal-oriented tasks, such as booking a flight from Chicago to London, where sequential actions lead directly to a predefined outcome without significant backtracking or exploration. These approaches address tasks with clear, predefined goals but overlook the challenge of aggregating information from multiple sources. 
In contrast, open-ended information-seeking tasks, such as investigating why the Indian education system lacks funding and infrastructure, require agents to explore multiple sources, consider diverse viewpoints, and determine when sufficient information has been gathered for a comprehensive answer.

Building agents for information-seeking tasks shifts the focus from linearized action sequences for goal completion to the quality and coverage of the aggregated information, highlighting a gap in current methods that do not consider such exploratory behaviors.
Specifically, we identify two critical limitations in current web agents: (1) \textit{Lack of Information Aggregation}: they cannot aggregate information from multiple webpages; and (2) \textit{Inability to Backtrack}: they are constrained to forward navigation, unable to revisit previous pages or explore alternative search results. These constraints hinder their effectiveness in information-seeking tasks that require iterative exploration. 

Motivated by the challenges, we introduce \textsc{Infogent}, a novel framework for information aggregation on the web which accomplishes the task using three specialized components: a \textit{Navigator} responsible for searching the web and identifying relevant websites, an \textit{Extractor} for identifying relevant information from the selected web pages, and an \textit{Aggregator} for selectively retaining the extracted information, and deciding what to include in the final aggregated output. To address the shortcomings of the current web agents, \textsc{Infogent} augments a task-completion agent with additional capabilities required to be an effective web navigator for information aggregation. Specifically, we introduce two key modifications: (1) \textit{Enhanced Action Set} that enables the navigator to backtrack and transfer control to other components when aggregation is to be performed; and (2) \textit{Feedback-Driven Navigation}, where navigator's decision-making process incorporates feedback from aggregator, ensuring that navigation strategies are dynamically informed by both the input query and the current state of information aggregation.

\textsc{Infogent} is modular, with a clear division of responsibilities between the three components, designed to operate effectively in real-world information aggregation settings. 
Specifically, we address two scenarios for accessing information from websites: \textit{Direct API-driven Access}, where agents are enabled access only to the textual web data extracted via APIs without visual interaction,  and \textit{Interactive Visual Access}, which requires agents to simulate visually-dependent human browsing to access web information, which can often be obstructed by paywalls, logins, or other necessary user interactions that can only be bypassed with visual context understanding. By evaluating on realistic, multi-website aggregation tasks--AssistantBench~\cite{yoran2024assistantbenchwebagentssolve}, FRAMES~\cite{krishna2024fact} and FanOutQA \citep{zhu-etal-2024-fanoutqa}--we demonstrate \textsc{Infogent}'s ability to effectively handle both information access settings.
%Our work evaluates \textsc{Infogent} on realistic, multi-document tasks, AssistantBench \citep{yoran2024assistantbenchwebagentssolve} and FanOutQA \citep{zhu-etal-2024-fanoutqa}, to demonstrate its ability to effectively handle multiple information sources while aggregating only the necessary information through the textual and visual understanding of the web. 
In summary, our contributions are as follows:
\begin{itemize}[noitemsep]
    \item We introduce \textsc{Infogent}, a novel modular and feedback-driven framework for web information aggregation through the use of three distinct components: Navigator, Extractor, and Aggregator (illustrated in Figure \ref{fig:infogent_overview}).
    %\item \textbf{Novel Problem Formulation} We define a novel problem of BFS-centric Web navigation task from the perspective of information aggregation. Unlike previous work that instructs Web agents to navigate within a fixed website with a fixed termination criteria in a linear sequence of action trajectory, our task demands the agents to explore outside the confines of a specific Web site, search, collect and aggregate information related to the given instruction.
    % \heng{all previous work limited to a fixed website? no DFS work is applied to multiple websites?}\revanth{Most previous web agent datasets have primarily been task-completion on single website. AssistantBench is a recent dataset on multi-website aggregation, which we have added evaluation on.} 
    \item We demonstrate that \textsc{Infogent} can be employed under both \textit{Direct API-Driven} Access and \textit{Interactive Visual Access} settings.
    %\item \textbf{Analysis of Programmatic and Non-Programmatic Web Navigation Agents} Our work provides empirical results and case studies on both the programmatic and non-programmatic approaches and highlight the importance of feedback mechanism in Web navigation for information aggregation. 
    \item On various web aggregation tasks, we empirically show that \textsc{Infogent} outperforms existing state-of-the-art multi-agent search frameworks and information-seeking web agents.
    %\item \textbf{Breadth-First Information Aggregation Benchmark} We also provide an information aggregation benchmark which is composed of complex instruction-following tasks prompted and extracted from InfoBench and WildBench to evaluate the Web navigation and information aggregation ability of Web agents.
\end{itemize}

% We identify several challenges in evaluating and designing web agents capable of performing BFS-centric web tasks:
% \begin{itemize}
%     \item Unlike previous work that builds the benchmark ontology upon different types of websites, BFS tasks usually do not confine to any single website, posing a challenge in task selection.
%     \item The open-ended nature of BFS tasks poses a challenge in the design of evaluation metrics.
%     \item The goal of BFS tasks is fundamentally different from DFS tasks, requiring a dedicated design of prompting and mechanisms for information aggregation.    
% \end{itemize}
\section{Related work}
\label{sec:related_work}
% \heng{add some related work about information aggregation, include your SmartBook paper}\zhenhailong{added}

\paragraph{Web Navigation with LLMs:}Web navigation agents were originally explored in simulated web environments~\citep{shi2017world} and \citep{liu2018reinforcement} which predominantly focused on completing goal-oriented tasks. The simulated environments came equipped with a range of task primitives such as selecting value from a drop down or entering text into an input box, which could be used to achieve the end goal. Subsequent work has focused on extending to more realistic settings~\citep{nakano2021webgpt}, such as WebShop~\citep{yao2022webshop} for e-commerce and RUSS~\citep{xu2021grounding} for web support. However, these efforts are still limited to a narrow set of domains and websites. 
In contrast, WebArena~\citep{zhou2023webarena} and Mind2Web~\citep{deng2024mind2web} were introduced as benchmarks for autonomous web agents that can generalize to a wide variety of tasks on real-world websites. Nevertheless, these approaches were still limited to predominantly language-guided agents, that solely relied on the text elements present within the website raw HTML. Follow-up works, such as VisualWebArena~\citep{koh2024visualwebarena}, SeeACT~\citep{seeact} and WebVoyager~\citep{he2024webvoyager}, use multimodal agents~\citep{achiam2023gpt, team2023gemini} that leverage screenshots of websites as input for identifying the appropriate HTML elements to act upon. The motivation is that raw HTML contents are too noisy, and context is often too long, while screenshots provide a less noisier view of the webpage. While these methods involve an autonomous agent solving the task using an initial instruction, more recently, WebLinx~\citep{lu2024weblinx} introduces the problem of \textit{conversational web navigation}, wherein the agent controls a real-world web browser and follows user instructions to solve tasks in a multi-turn dialogue fashion.

\paragraph{Web Information Aggregation:}Recently, there has been a growing interest for more complex information aggregation tasks, which have been studied independently within the Information Extraction field~\citep{reddy2023smartbook}. In the context of Web Agents, information aggregation requires broader exploration and backtracking to effectively generate the answer.
MindSearch \citep{chen2024mindsearchmimickinghumanminds} explores this, modeling the task as an iterative graph construction. AssistantBench \citep{yoran2024assistantbenchwebagentssolve} enhances SeeAct with the go back action and a planning module, and tackles time consuming tasks on the web. In this work, we propose \textsc{Infogent}, a modular framework featuring specialized aggregation and feedback modules that achieves state-of-the-art performance in both Direct API-driven Access and Interactive Visual Access scenarios.
%\zhenhailong{add our distinctive contribution here compared with AssistantBench/MindSearch}

% \begin{comment}
%     \begin{itemize}
%         \item WebLinx \citepp{lu2024weblinx}
%         \item WebArena \citepp{zhou2023webarena}
%         \item WebGPT \citepp{nakano2021webgpt}
%         \item Mind2Web \citepp{deng2024mind2web}
%         \item AutoGPT \citepp{}
%     \end{itemize}
% \end{comment}
    
\section{Information Aggregation Task}
\label{sec:info_agg_task}
%We conceptualize information aggregation for a given user query as an iterative process that involves (i) identifying relevant websites, (ii) gathering pertinent information within them, and repeating (i) and (ii) until sufficient information is aggregated. A key aspect of this process is actively keeping track of the aggregated information, which guides the system in identifying what information needs to be sought in subsequent steps. This approach ensures comprehensiveness in the aggregated data while avoiding redundancy. The success of the aggregation process is evaluated based on the quality, diversity, and sufficiency of the final collected information. The accessibility of web information varies. Some are easily obtainable through APIs provided by search engines or platforms. For example, information about the ``Billboard Top 100 songs'' can be retrieved by  by scraping the relevant Wikipedia page titled ``Billboard Hot 100''. However, there are instances where the desired information is not readily accessible through basic search APIs. For instance, salary data for specific job titles or fundraising details for startups is often hosted on sites like Glassdoor and Crunchbase, which require user authentication or subscriptions. Consequently, paywalled or login-protected information cannot be easily accessed via standard URL scraping techniques.

We conceptualize information aggregation for a query as an iterative process involving identifying relevant websites and gathering pertinent information within them, repeated until sufficient data is collected. Actively tracking the aggregated information guides subsequent searches, ensuring comprehensiveness while avoiding redundancy. The success of the process is dependent on the quality and diversity of the collected information. 

We note that accessibility of web information varies significantly. Some data is easily obtainable through APIs or by scraping web pages (e.g.,  retrieving ``Billboard Top 100 songs'' from Wikipedia ). However, other information, such as salary data on Glassdoor, is not directly accessible due to paywalls, or other restrictions. Therefore, we categorize information aggregation tasks into two settings based on the type of access: \textit{Direct API-Driven Access} and \textit{Interactive Visual Access}. 

The former involves retrieving data via APIs or automated tools without interacting with the website, making it efficient when APIs are available. In contrast, Interactive Visual Access requires simulating human browsing to retrieve information from screenshots of webpages that prohibit automatic scraping. 
We hypothesize that these two approaches together encompass a wide range of practical scenarios for information aggregation, and any comprehensive solution should handle both paradigms. Moreover, while we primarily focus on web-based aggregation, the concept of Interactive Visual Access extends to other desktop and mobile applications, such as Slack or iMessage, where API access is restricted and visual interaction is necessary~\cite{ge2023llm, kapoor2024omniact}.

\begin{figure*}[t]
    \centering
    \includegraphics[width=1.0\linewidth]{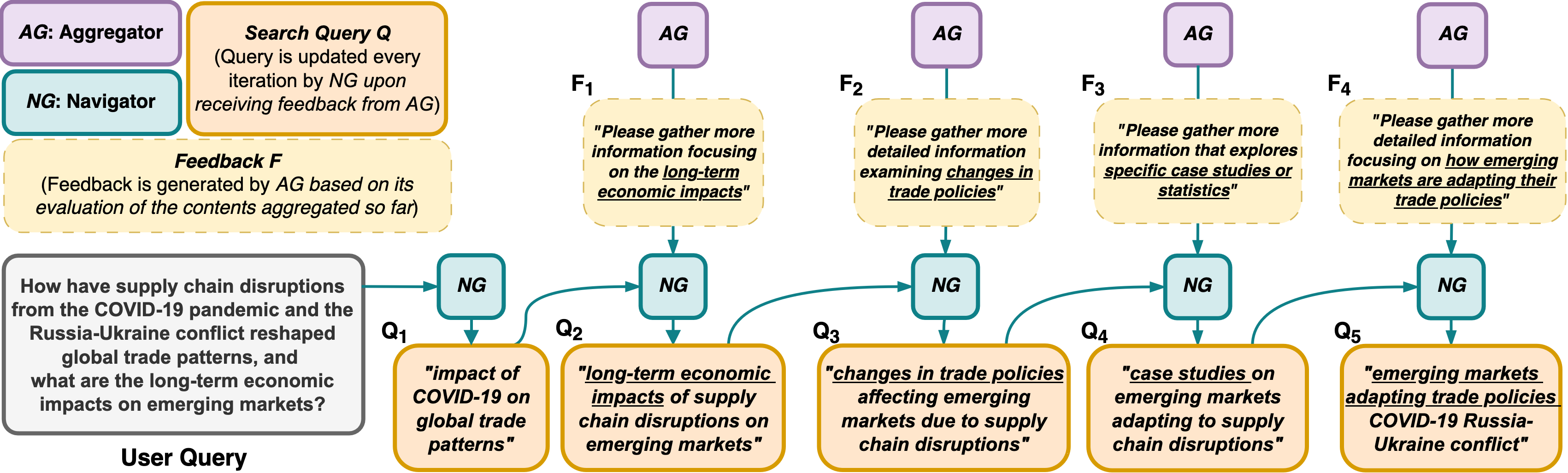}
    \vspace{-1.5em}
    \caption{A working example of \textsc{Infogent}. $\mathcal{NG}$ iteratively generates an updated query given feedback from $\mathcal{AG}$.}
    \label{fig:programmatic_example}
    \vspace{-1em}
\end{figure*}
\section{\textsc{Infogent}}
\label{sec:infogent}
% We present \textsc{Infogent}, our proposed framework for information aggregation, structured around three core components: a \textit{Navigator}, responsible for identifying relevant websites; the \textit{Extractor}, which extracts specific information from these sites based on the user query; and the \textit{Aggregator} which monitors the aggregated content and determines whether newly extracted information should be included.

%\textsc{Infogent} adopts a clear division of responsibility between the three components. 

\begin{algorithm}[t]
    \small
    \caption{\small Information Aggregation with \textsc{Infogent}}
    \label{alg:info_agg}
    %\linespread{1.08}
    \SetAlgoLined
    \DontPrintSemicolon  % Removes semicolons at the end of each line
    \KwIn{$\mathcal{T}$: User Task, $K$: Max websites to extract, $N$: Max time steps}
    \KwOut{$\mathcal{S}$: Aggregated information stack}
    
    %$\mathcal{T} \gets \text{``Input User Query''}$, 
    %$N \gets \text{"Max Websites to Extract"}$, 
    %$\mathcal{T}_{flag} \gets \text{False}$ \tcp*{Termination Flag}
    $\mathcal{W}_0 \gets \text{"Search Home"}$ \tcp*{Starting Webpage}
    %$\mathcal{P} \gets \text{"None"}$ \tcp*{Current Extracted Information}
    $\mathcal{S}_0 \gets \text{"Empty Stack"}$ \tcp*{Information Stack}
    $\mathcal{F} \gets \text{"None"}$ \tcp*{Aggregator Feedback}
    $k \gets 0; t \gets 0$ \tcp*{Iteration \& Action Counter}     
    % $k \gets 0$ \tcp*{Iteration Counter}    
    % $t \gets 0$ \tcp*{Action Counter}   
    \While{$a_t != \texttt{TERMINATE}$ \textbf{and} $k < K$ \textbf{and} $t < N$}{
        $a_{t+1} = \mathcal{NG}(\mathcal{W}_t, \mathcal{T, F,} \{a_1, a_2, \dots, a_t\})$\;
        \If{$a_{t+1} = \texttt{AGGREGATE}$}{
            $\mathcal{P} \gets \mathcal{ET}(\mathcal{W}_t, \mathcal{T}, \mathcal{F})$ \tcp*{Extract Info.}
            $\mathcal{S}_{k+1}, \mathcal{F} \gets \mathcal{AG}(\mathcal{S}_k, \mathcal{P}, \mathcal{T})$ \tcp*{Update $\mathcal{S}_k$}
            $k \gets k + 1$ \tcp*{Update Counter}
            $W_{t+1} \gets W_t$ 
        }
        \Else{
            $\mathcal{W}_{t+1}$ = Act($\mathcal{W}_t$, $a_{t+1}$)   \tcp*{Make Action}
        }
        $t \gets t + 1$         \tcp*{Update Counter}
        
    }
    %\Return $\mathcal{S}_N$
\end{algorithm}

\textsc{Infogent}, as illustrated in Fig. \ref{fig:infogent_overview}, consists of three core components: A \textit{Navigator} $\mathcal{NG}$, an \textit{Extractor} $\mathcal{ET}$, and an \textit{Aggregator} $\mathcal{AG}$. Given an information-seeking query, the Navigator $\mathcal{NG}$ initiates the process by searching the web for relevant sources. Upon identifying a suitable webpage, the Extractor $\mathcal{ET}$ takes over the control, which extracts relevant content and forwards it to the Aggregator $\mathcal{AG}$. $\mathcal{AG}$ evaluates this content with respect to the information aggregated so far and decides whether to include it. Importantly, $\mathcal{AG}$ provides feedback to $\mathcal{NG}$ about gaps in the aggregated information, guiding subsequent searches to address deficiencies. $\mathcal{NG}$ lacks direct access to the aggregated information, thereby relies on $\mathcal{AG}$'s feedback for directions in subsequent iterations. This iterative process continues until $\mathcal{AG}$ determines that sufficient information has been gathered and instructs $\mathcal{NG}$ to halt. Thus, \textsc{Infogent} employs a modular, feedback-driven approach to information aggregation, making it suitable for complex queries requiring diverse sources. Fig.  \ref{fig:programmatic_example} illustrates the feedback-driven navigation with example.

Let's denote the action space of $\mathcal{NG}$ as $\mathcal{A}$, the task at hand as $\mathcal{T}$, the Aggregator feedback as $\mathcal{F}$ and the current website under consideration as $\mathcal{W}$. Further, there is a stack $\mathcal{S}$ of diverse information aggregated in the form of a list of paragraphs, which is returned upon task completion. %$\mathcal{O}$ and $\mathcal{A}$ for both our approaches are detailed later. \textcolor{blue}{refer where it has been used}
$\mathcal{NG}$ is responsible for navigating the internet to identify relevant web pages. Formally, at time step $t$, for a given website $\mathcal{W}$, $\mathcal{NG}$ samples an action $a_t \in \mathcal{A}$ from its action space (shown in Table \ref{tab:navigator_action}), which varies depending on the information access setting. 
\begin{equation*}
 a_t = \mathcal{NG}(\mathcal{W, T, F,} \{a_1, a_2, \dots, a_{t-1}\})   
\end{equation*}

If the action $a_t$ is \texttt{AGGREGATE}, $\mathcal{ET}$
extracts relevant information from $\mathcal{W}$ for the task $\mathcal{T}$, to provide a list of passages $\mathcal{P} = \mathcal{ET}(\mathcal{W, T, F})$.
%\(\mathcal{I}(a_t)\) \text{ is the identity function applied to } \(a_t\) \text{ ( indicating it is \texttt{INFORMATION AGGREGATION} ).} 
%\dilek {This comes out a bit confusing, as you haven't mentioned other types of actions yet. What other actions could there be? Also, AG outputs T\_flag in the algorithm, it would help to say that it is the flag to halt the process in the next paragraph. Also, explain what Tflag is in your algo, its confusing since its not explained before. }
%\revanth{I have added reference to other actions for NG, and we have removed the termination flag since it's being halted using the TERMINATE navigator action based on feedback.}
$\mathcal{AG}$ then evaluates the relevance of $\mathcal{P}$ according to the current information stack $\mathcal{S}$ and the task $\mathcal{T}$, updates $\mathcal{S}$. and returns natural language feedback $\mathcal{F}$, to guide $\mathcal{NG}$'s subsequent actions. Using $\mathcal{F}$, $\mathcal{AG}$ can also instruct $\mathcal{NG}$ to finish the process once sufficient information has been aggregated. Algo. \ref{alg:info_agg} shows a schematic of \textsc{Infogent}'s working process.

\textsc{Infogent}'s modular architecture is optimized for information aggregation and enhances adaptability across diverse scenarios by dividing responsibilities among distinct components. $\mathcal{NG}$ and $\mathcal{ET}$ can utilize either language-only or multimodal models, depending on the nature of web information access discussed in \S{\ref{sec:info_agg_task}}. Given our primary focus on textual information aggregation, both access types employ the same aggregator component. Further details on $\mathcal{NG}$, $\mathcal{ET}$, and $\mathcal{AG}$ follow.

\subsection{Navigator $\mathcal{NG}$}
\label{sec:navigator}

Recent studies~\cite{yang2023autogpt, wang2024survey} have demonstrated the capabilities of LLMs and LMMs to autonomously plan and execute sequences of thoughts and actions~\cite{yao2023react} based on a high-level directive. Building on this capability, we conceptualize the navigator as an autonomous agent tasked with exploring the web to identify relevant websites. The action space available to the navigator agent, shown in Table \ref{tab:navigator_action}, depends on the information access setting.  Specifically, under the Direct API-Driven Access setting, 
%\dilek{Is the setting determined manually or automatically? Also, is it either/or, or can both be triggered for aggregating information?}\revanth{the setting is determined manually, and it is either/or} 
\textsc{Infogent} employs a tool-based LLM agent~\cite{yang2023autogpt} as the Navigator, which leverages a search API as a tool. Conversely, in the Interactive Visual Access setting, a multimodal web navigation agent~\cite{zheng2024seeact} is utilized to interact with a real-world browser and access relevant content within the webpages. The Navigator here simulates human-like browsing behavior, allowing the agent to navigate through web interfaces that may not be accessible via APIs alone.

\begin{table}[t]
\small \centering
\begin{subtable}[t]{0.45\textwidth}
\centering
\caption{Direct API-Driven Access}
\label{tab:direct_api_action}
\def\arraystretch{1.2}
\begin{tabular}{ll}
\toprule
\textbf{Action}      & \textbf{Description}                                           \\ \midrule
\texttt{SEARCH} (query)        & Return top-5 (url, snippet) pairs                           \\
\texttt{AGGREGATE} ($\mathcal{W}$)       & calls \(\mathcal{ET}\) and \(\mathcal{AG}\) in sequence        \\ 
\texttt{TERMINATE}           & Terminate navigation                                           \\\bottomrule
\end{tabular}
\end{subtable}%
%\hspace{0.1\textwidth} % Adjust the space between the subtables

\begin{subtable}[t]{0.45\textwidth}
\centering
\caption{Interactive Visual Access}
\label{tab:interactive_visual_action}
\def\arraystretch{1.2}
\begin{tabular}{ll}
\toprule
\textbf{Action}      & \textbf{Description}                                           \\ \midrule
\texttt{CLICK} (element)              & element.click()                                                \\
\texttt{SELECT} (element)             & element.select()                                               \\
\texttt{TYPE} (element, text)               & Type text in selected element                                  \\
\texttt{PRESS ENTER}         & Press enter                                                    \\
\texttt{GO BACK}             & Go back to previous page                                       \\
%NONE                & null                                                           \\
\texttt{AGGREGATE} ($\mathcal{W}$)          & calls \(\mathcal{ET}\) and \(\mathcal{AG}\) in sequence        \\ 
\texttt{TERMINATE}           & Terminate navigation                                           \\\bottomrule
\end{tabular}
\end{subtable}
\vspace{-0.5em}
\caption{Action space $\mathcal{A}$ of the Navigator $\mathcal{NG}$.}
\label{tab:navigator_action}
\vspace{-1em}
\end{table}

\subsubsection{Direct API-Driven Access}
\label{sec:programmatic_method}
In this setting, web information can be accessed by querying a search API, which returns a list of relevant urls; the corresponding website content can be retrieved using automated scraping tools. In this context, $\mathcal{NG}$ is an autonomous agent~\cite{yang2023autogpt}, based on the ReACT framework~\cite{yao2023react}, which combines chain-of-thought ~\citep{wei2022chain} with tool calls to generate sequence of thought and actions. 

The action space $\mathcal{A}$ of $\mathcal{NG}$ under this setting (shown in Table \ref{tab:direct_api_action}), consists of two tools, namely \texttt{SEARCH} and \texttt{AGGREGATE}. Given the user task, $\mathcal{NG}$ employs \texttt{SEARCH}\footnote{Our experiments use Google Search as the search API.} with an appropriate query, resulting in a set of URLs accompanied by brief descriptive snippets. $\mathcal{NG}$ then chooses a relevant URL from this set to invoke the \texttt{AGGREGATE} tool, which encompasses both $\mathcal{ET}$ and $\mathcal{AG}$. $\mathcal{ET}$ first scrapes the URL and extracts relevant content $\mathcal{P}$. Next, $\mathcal{AG}$ updates $\mathcal{S}$ using $\mathcal{P}$, and returns textual feedback $\mathcal{F}$. Based on $\mathcal{F}$, $\mathcal{NG}$ adjusts its strategy accordingly: if the extracted content is affirmed as relevant and useful, it continues to explore additional websites from the initial search results; if the content is deemed irrelevant or redundant, it initiates a new search with a revised query informed by $\mathcal{AG}$'s feedback.

\subsubsection{Interactive Visual Access}
\label{sec:programmatic_method}

Under this setting, information cannot be directly scraped, meaning $\mathcal{NG}$ needs to explore the web in a manner similar to human interactions with a browser. Recent work~\citep{he2024webvoyager, seeact} has demonstrated promising results in leveraging powerful Large Multimodal Models (LMMs)~\citep{openai2023gpt4v} for web navigation. The navigator here is based on SeeAct~\cite{seeact}, a task-completion agent, capable of finishing web tasks by planning and executing interactive actions on webpages by utilizing screenshots and candidate HTML elements. SeeAct first performs \textit{Action Generation} to create natural language descriptions of the necessary actions to accomplish a task (e.g., ``Click on search button''). Subsequently, it engages in \textit{Action Grounding} to identify appropriate HTML elements (e.g., ``[input] Departure City'') and determines the corresponding operations (such as \texttt{CLICK}, \texttt{TYPE} etc.) to execute. For more details on SeeAct, please refer to~\citet{seeact}. 

%However, we see two critical limitations in existing web agents: (1) \textit{Lack of Information Aggregation:} They primarily focus on task completion, such as booking a flight or purchasing an item, without the ability to aggregate information as they navigate through multiple webpages. (2) \textit{Inability to Backtrack:} Most of these agents are constrained to forward navigation, lacking the capability to backtrack or explore other search results. The ability to \textit{go back} and \textit{explore} alternative sources is crucial for information-seeking tasks that usually require iterative exploration. To address these shortcomings, \textsc{Infogent} augments SeeAct, a task-completion agent, with additional capabilities required to be an effective web navigator for information aggregation. Specifically, we introduce two key modifications: (1) \textit{Enhanced Action Set:} We incorporate \texttt{GO BACK} and \texttt{AGGREGATE} actions, enabling the agent to perform backtracking and to transfer control to the Extractor respectively; (2) \textit{Feedback-Driven Navigation:} We modify the Action Generation procedure to also condition on feedback from the Aggregator. This feedback mechanism ensures that the navigation strategy is not only dependent on the input user query, but also dynamically informed by the current state of information aggregation.

We augment SeeAct with additional capabilities required to be an effective web navigator for information aggregation. We add \texttt{GO BACK} and \texttt{AGGREGATE} actions, enabling the agent to perform backtracking and to transfer control to $\mathcal{ET}$ respectively. The full list of actions is provided in Table \ref{tab:interactive_visual_action}. Further, we modify the Action Generation procedure to also condition on the textual feedback $\mathcal{F}$ from $\mathcal{AG}$. The navigation begins from the search engine home page, with $\mathcal{NG}$ leveraging the \texttt{CLICK}, \texttt{SELECT}, \texttt{TYPE}, and \texttt{PRESS ENTER} actions to get the search results and explore the web pages further. The $\texttt{AGGREGATE}$ action is used to invoke $\mathcal{ET}$ and $\mathcal{AG}$ when the webpage is deemed relevant. Subsequently, based on the feedback $\mathcal{F}$, $\mathcal{NG}$ leverages the \texttt{GO BACK} action to retrace its steps to explore other search results, or instead perform another search using a different revised query.

%The enhanced Interactive Visual Navigator $\mathcal{NG}$ has an action space as detailed in \ref{tab:action_space_non_programmatic}. Further the observation space for it looks like $\mathcal{O}$ = [$\mathcal{F}$, $\mathcal{T}$, Screenshot of current url]. 

% \begin{table}[t]
%   \small \centering
%   \caption{\small Action space of Interactive-Visual $\mathcal{NG}$}
%   \label{tab:action_space_non_programmatic}
%   \begin{tabular}{ll}
%     \toprule
%     \textbf{Action}                   & \textbf{Description}       \\
%     \midrule
%     \texttt{CLICK}           & element.click()
%            \\
%     \texttt{SELECT}                & element.select()     \\
%     \texttt{TYPE}                & type text in selected element     \\
%     \texttt{PRESS ENTER}                & press enter     \\
%     \texttt{GO BACK}                & Go back to previous page     \\
%     \texttt{TERMINATE}                & terminate navigation     \\
%     \texttt{NONE}                & null     \\
%     \texttt{AGGREGATE}                & calls $\mathcal{ET}$ and $\mathcal{AG}$ in sequence     \\
%     \bottomrule
%   \end{tabular}
% \end{table}

%\vspace{-1em}
 
\subsection{Extractor $\mathcal{ET}$}
\label{sec:extractor}

%Once the Navigator selects a relevant website, the Extractor's role is to identify and extract up to $k$ paragraphs relevant to the user task. Given that webpages are often considerably long, employing a smaller, more cost-efficient model for content processing is practical. We prefer extraction over summarization of relevant content for two primary reasons. First, smaller models may produce summaries of questionable quality due to their limited capacity. Second, such models are prone to hallucination, generating information that is not present in the source text. Direct extraction ensures accurate attribution for downstream tasks, maintaining the reliability of aggregated data. 
Once $\mathcal{NG}$ selects a relevant website, $\mathcal{ET}$ identifies and extracts up to $k$ relevant paragraphs for the task. Since webpages are often lengthy, using a smaller, cost-efficient model for content processing is more practical. Extraction is favored over summarization for two key reasons: smaller models tend to produce low-quality summaries due to limited capacity, and they are prone to hallucination, introducing information not present in the source. Direct extraction ensures accurate attribution and maintains reliability of the aggregated data.

%\heng{I wonder whether paragraph-level is too coarse grained for extracting specific information like the answers for your example in Figure 2; and for the ADD and REPLACE actions later, how do you decide which information is related?} \revanth{The system's output is intended to be passed onto a downstream LLM for response generation. Hence, to keep recall high, the goal is to ensure that the specific information needed is contained in these paragraphs. With regards to the ADD and REPLACE actions, the aggregator is prompted to add / replace the paragraph if the information within it is relevant to the task and non-redundant with respect what was already aggregated.}

%Under the Interactive Visual Access setting, since the website content cannot be directly scraped due to access restrictions, the Extractor scrolls through the webpage from top to bottom, capturing multiple screenshots. These screenshots are then fed to a multimodal model~\cite{openai2023gpt4v} to output the relevant paragraphs present on the page. This enables the extraction of information from web interfaces that is otherwise inaccessible through traditional scraping techniques. We refer the reader to Table \ref{table:programmatic_prompts} in the appendix for detailed prompts.

In the Direct API-Driven Access setting, given a website URL, $\mathcal{ET}$ scrapes the content and feeds it into an LLM, which is prompted to identify the relevant paragraphs based on the user's task.  In contrast, under the Interactive Visual Access setting, where website content cannot be directly scraped due to access restrictions, $\mathcal{ET}$ navigates the webpage by scrolling from top to bottom, capturing multiple screenshots. These screenshots are then processed by a multimodal model~\cite{openai2023gpt4v}, which identifies and extracts the relevant paragraphs. This approach facilitates extraction from web interfaces that are otherwise inaccessible through traditional scraping techniques. For detailed prompts, refer to Table \ref{table:programmatic_prompts} in the Appendix.

\subsection{Aggregator $\mathcal{AG}$}
\label{sec:aggregator}
Given the content extracted by $\mathcal{ET}$, presented as a list of paragraphs $\mathcal{P}$, $\mathcal{AG}$'s task is to determine whether to incorporate any of the paragraphs into the aggregated information stack $\mathcal{S}$. For each passage $p_i$ in $\mathcal{P}$, $\mathcal{AG}$ can choose to either add $p_i$ as a new item (\texttt{ADD($p_i$)}), replace an existing item $s_j$ in $\mathcal{S}_t$ with $p_i$ (\texttt{REPLACE($s_j$, $p_i$)}) or just ignore $p_i$ if it is irrelevant or redundant. This decision-making process is achieved by prompting an LLM, with detailed prompts in Table \ref{table:programmatic_prompts} in the Appendix.
Furthermore, $\mathcal{AG}$ provides textual feedback $\mathcal{F}$ to $\mathcal{NG}$ regarding what information to seek next, which guides the $\mathcal{NG}$'s subsequent actions by highlighting information gaps in $\mathcal{S}$. By incorporating a feedback-driven interaction between $\mathcal{AG}$ and $\mathcal{NG}$, \textsc{Infogent} ensures the information-seeking process is adaptive to the aggregated information.

%If none of the passages are relevant to the task, or if the information is already available in the aggregated data, the Aggregator opts to not include any of the extracted information.  

%on what information to look for, which helps guide the navigator on how to proceed next. 

% 

% Our framework isolates \heng{not sure what you mean by isolate?} the aggregated data within the aggregator to promote modularity. Specifically, this design choice facilitates exploration of alternate approaches for navigation, such as using vision-language models, in setting where text-only models for navigation are not enough. \heng{this paragraph is a bit vague and hard to follow}

\section{Experiments}

We test \textsc{Infogent}'s ability to address complex queries that require accumulating information over multiple webpages. Evaluation is based on the final answer generated by the downstream LLM, leveraging the information aggregated by \textsc{Infogent}. We consider evaluation separately for Direct API-Driven access and Interactive Visual Access. 

%To evaluate our proposed framework, we consider the task of multi-hop Web Question Answering (QA), which requires accumulating information over multiple webpages to be able to answer the question. Specifically, we evaluate on FanOutQA~\citep{zhu-etal-2024-fanoutqa}, which comprises multi-hop questions that require finding information on the web about a multiple entities (for eg. \textit{What is the population of the five smallest countries, by area, located in the largest continent?}). The dynamic nature of web-based QA means the answers to the questions can change as the web pages get updated. Hence, following ~\cite{zhu-etal-2024-fanoutqa}, we only consider English Wikipedia pages to answer the questions. This choice is due to the versioning in Wikipedia enabling retrieving the page content corresponding to when the dataset was annotated, meaning the answers are consistent over time. We evaluate \textsc{Infogent} under both Direct API-Driven Access (in \S{\ref{sec:api_driven_access}}) and Interactive Visual Access (in \S{\ref{sec:interactive_visual_access}}) settings on the FanOutQA benchmark. 

\subsection{Direct API-Driven Access}
\label{sec:api_driven_access}

 Here, we employ a tool-based LLM as $\mathcal{NG}$, built upon AutoGPT. To mitigate issues arising from the dynamic and potentially conflicting information on the web, we restrict our search to Wikipedia pages, following prior work~\citet{zhu-etal-2024-fanoutqa}.

\subsubsection{Setup}
\label{sec:direct_setup}

\paragraph{Datasets and Metrics:}  We evaluate our method on the FanOutQA~\cite{zhu-etal-2024-fanoutqa} and FRAMES~\cite{krishna2024fact} datasets, both comprising complex queries that require accumulating information from multiple webpages. FanOutQA includes 310 multi-hop questions involving multiple entities (for e.g. \textit{What is the population of the five smallest countries by GDP in Europe?}). FRAMES contains complex questions requiring various reasoning types: numerical (counting, comparisons, calculations), tabular (using statistics from tables or infoboxes), constraints (multiple conditions leading to a unique answer), temporal (timeline reasoning) and post-processing (specific steps after gathering all necessary facts). Excluding numerical questions--whose performance depended significantly on the final answering LLM rather than the aggregation approach--we retained 531 examples. We use the official evaluation metrics for both datasets:  FanOutQA employs string accuracy and ROUGE~\cite{chin2004rouge}, while FRAMES uses language model to assess whether the generated output matches the gold answer, utilizing the prompt shown in Table~\ref{table:evaluation_task} in the Appendix.

\paragraph{Baselines:} We compare \textsc{Infogent} with MindSearch~\cite{chen2024mindsearchmimickinghumanminds}, a multi-agent search framework involving a planner and a searcher. MindSearch models information seeking as a dynamic graph construction process via code-driven decomposition of the user query into atomic sub-questions represented as nodes. It then iteratively builds the graph for the subsequent steps, based on answers to the sub-questions. The output is then passed to a downstream LLM for answer generation, similar to \textsc{Infogent}. We also include a closed-book model as a baseline. All approaches employ GPT-4o-mini as the underlying LLM.

\begin{table}[t]
\centering
\small
\setlength{\tabcolsep}{2pt} 
\begin{tabular}{llcccc}
\toprule
    \textbf{Approach} & \textbf{All} & \textbf{Tabular} & \textbf{Temporal} & \textbf{Constr.} & \textbf{Process} \\
\midrule
    Closed-Book & 23.5 & 16.4 & 19.9 & 22.7 & 11.6 \\
    MindSearch & 46.3 & 41.4 & \textbf{46.6} & 47.5 & 30.0  \\
\midrule
    \textsc{Infogent} & \textbf{53.3} & 
\textbf{45.7} & 43.8 & \textbf{55.2} & \textbf{46.5} \\
\bottomrule
\end{tabular}
\vspace{-0.5em}
\caption{Results (in \%) on the Frames dataset for queries with different reasoning types under Direct API-Driven Access setting. Constr. corresponds to Constraints.} 
\label{tab:results_api_driven_frames}
\end{table}

\begin{table}[t]
\centering
\small
\begin{tabular}{lcccc}
\toprule
    \textbf{Approach} & \textbf{Acc.} & \textbf{R-1} & \textbf{R-2} & \textbf{R-L} \\
\midrule
    Closed-Book & 46.6 & 44.5 & 24.2 & 38.2\\
    MindSearch & 47.3 & 49.3 & 28.4 & 44.2 \\
\midrule
    \textsc{Infogent} & \textbf{51.1} & \textbf{53.3} & \textbf{33.0} & \textbf{48.5} \\
\bottomrule
\end{tabular}
\vspace{-0.5em}
\caption{Results (in \%) on the FanoutQA dev set under the Direct API-Driven Access setting.}
\label{tab:results_api_driven}
\vspace{-0.5em}
\end{table}

\subsubsection{Results}
\label{sec:direct_setup}

Table \ref{tab:results_api_driven_frames} reports results on FRAMES across different reasoning types. Low performance of the closed-book approach highlights the complexity and recency of the questions.  \textsc{Infogent} significantly outperforms MindSearch on most reasoning types; however, on temporal reasoning, MindSearch performs better, likely due to its code-driven planning in graph construction. Table \ref{tab:results_api_driven} presents results on FanOutQA. Both \textsc{Infogent} and MindSearch outperform the closed-book method, demonstrating the benefit of web search, with \textsc{Infogent} consistently surpassing MindSearch. The relatively high performance of the closed-book model may be due to the dataset's release date (Nov 2023) being close to the LLM's knowledge cutoff (Oct 2023), suggesting that the LLM's parametric knowledge might already contain the required facts.
% \dilek{Closedbook seems very high as well. Can you discuss why?} 

%We also evaluate \textsc{Infogent} on the Frames \cite{krishna2024fact} benchmark. We filtered the queries tagged \textit{numerical}, since it requires reading comprehension on top of information seeking. Our results are recorded in Table \ref{tab:results_api_driven_frames}.

\subsection{Interactive Visual Access}
\label{sec:interactive_visual_access}
Our Navigator $\mathcal{NG}$ in this setting uses the same web browser simulation tool as in \textsc{SeeAct}~\cite{zheng2024seeact}, built on top of Playwright. The navigator initiates search from the Google homepage.

\subsubsection{Setup}
\label{sec:interactive_setup}
\paragraph{Datasets and Metrics:} We use AssistantBench~\cite{yoran2024assistantbenchwebagentssolve}, a dataset for evaluating web agents on time-consuming online information-seeking tasks, such as monitoring real estate markets or locating relevant nearby businesses. It comprises 214 realistic tasks (33 dev and 181 test) that require interacting with multiple websites. To assess performance on information-dense websites (Wikipedia) under the interactive visual access setting, we use a human-curated subset of FanOutQA released by \citet{yoran2024assistantbenchwebagentssolve}, containing 31 queries with updated answers where closed-book models fail. Following~\citet{yoran2024assistantbenchwebagentssolve}, answer accuracy is the eval metric for both datasets.

\begin{table}[t]
\centering
\small
\begin{tabular}{llccc}
\toprule
\textbf{Type} & \textbf{Approach} & \textbf{Model} & \textbf{Dev} & \textbf{Test} \\
\midrule
\multirow{2}{*}{RAG} & RALM-Inst & GPT-4T & 15.5 & 11.7 \\
                                         & RALM-1S & GPT-4T & 13.6 & 10.6 \\
\midrule
\multirow{2}{*}{Web Agent} & \textsc{SeeAct}& GPT-4T &  0.0 & 4.2 \\
                            & SPA & GPT-4T & 12.7 & 11.0 \\
\midrule
\multirow{2}{*}{Web Agent}      & \multirow{2}{*}{\textsc{Infogent}} & GPT-4o  & 19.2 & \textbf{15.3}\\ 
       & & GPT-4T  & \textbf{22.0} & --\\

\bottomrule
\end{tabular}
\vspace{-0.5em}
\caption{Accuracy (in \%) on AssistantBench in Interactive Visual Access Setting.  Baseline numbers are taken from~\citet{yoran2024assistantbenchwebagentssolve}. } 
\label{tab:results_assistantbench}
\vspace{-0.7em}
\end{table}
%\vspace{-1em}

\paragraph{Baselines:} Baselines are same as in in~\citet{yoran2024assistantbenchwebagentssolve}. RALM-Inst and RALM-1S are zero and one-shot versions of a retrieval-augmented LM that is prompted to use Google Search as a tool~\cite{yao2023react}. For web-agent baselines, we consider \textsc{SeeAct}~\citep{seeact}, designed for web task-completion. Our primary comparison is with SPA (See-\textit{Plan}-Act)~\cite{yoran2024assistantbenchwebagentssolve}, which extends \textsc{SeeAct} for information-seeking tasks by incorporating planning and memory modules for information transfer between steps. 

\subsubsection{Results}
\label{sec:interactive_results}

Table \ref{tab:results_assistantbench} presents results on AssistantBench, where \textsc{Infogent} outperforms SPA by 6.5\% and 4.5\% on the dev and test sets respectively, even when using the smaller GPT-4o as backbone. Due to cost considerations, we report results on dev set with GPT-4T, observing a performance gain of 9.3\% over SPA. The poor performance of \textsc{SeeAct} confirms that task-completion web agents struggle with web information-seeking tasks. 
Table \ref{tab:results_interactive_visual} summarizes our results on FanOutQA, where \textsc{Infogent} improves upon the SPA baseline by 19\%. Since navigator is often the point of failure in web tasks, the Extractor and Aggregator in \textsc{Infogent} lower the burden on the Navigator, unlike in SPA, where a single agent handles navigation, planning and memory management. Thus, \textsc{Infogent}'s modular approach, with a clear division of responsibility between the components, contributes to its superior performance.

%To evaluate our method on information seeking tasks on the web we use the existing FANOUTQA \cite{zhu-etal-2024-fanoutqa} benchmark. It involves tasks that requires information aggregation from wikipedia pages and answering multi-hop questions on top of the documents aggregated. 
%\textbf{FanOutQA:} Table \ref{tab:results_interactive_visual} summarizes our results. We improve upon the existing SPA web-agent baseline by 19\%. 

% \begin{table}[ht]
% \centering
% \small
% \begin{tabular}{p{6em}lcc}
% \toprule
% \textbf{Type} & \textbf{Model} & \textbf{Acc. \%} & \textbf{Ans. \%} \\
% \midrule
% \multirow{2}{*}{\parbox{6em}{Closed-book LMs}} & CB-Inst & 16.3 & 53.5 \\
%                                  & CB-1S & 22.2 & 89.1 \\
% \midrule
% \multirow{2}{*}{RAG} & RALM-Inst & 11.7 & 59.8 \\
%                                          & RALM-1S & 10.6 & 48.3 \\
% \midrule
% \multirow{2}{*}{Web Agent} & \textsc{SeeAct}& 4.2 & 20.0 \\
%                             & SPA & 11.0 & 38.8 \\
% \midrule
% Web Agent       & \textsc{\parbox{6em}{\textsc{Infogent} \\ +GPT4O}}  & 15.3 & 91.9\\ 

% \bottomrule
% \end{tabular}
% \vspace{0.5em}
% \caption{Results on the Assistantbench testset under the Interactive Visual Access Setting. All methods use GPT-4-Turbo as the underlying model (unless otherwise mentioned), and numbers for the baselines are directly taken from~\citep{yoran2024assistantbenchwebagentssolve}.}
% \label{tab:results_assistantbench_dev}
% \end{table}

\begin{table}[t]
\centering
\small
\begin{tabular}{p{6em}lcc}
\toprule
\textbf{Type} & \textbf{Approach} & \textbf{Acc. \% } \\
\midrule
%\multirow{2}{*}{\parbox{6em}{Closed-book LMs}} & CB-Inst & 34.8 & 93.5 \\
 %                                & CB-1S & 40.9 & 100 \\
%\midrule
\multirow{2}{*}{RAG} & RALM-Inst & 9.6  \\
                                         & RALM-1S & 27.3 \\
\midrule
\multirow{2}{*}{Web Agent} & \textsc{SeeAct}& 7.5  \\
                            & SPA & 30.0  \\
                            \midrule
                   Web Agent         & \textsc{Infogent} & \textbf{49.0}\\ 

\bottomrule
\end{tabular}
\vspace{-0.5em}
\caption{Results on the human-curated subset of FanOutQA under the Interactive Visual Access Setting. All methods use GPT-4T as underlying model, and numbers for the baselines are taken from~\citet{yoran2024assistantbenchwebagentssolve}.}
\label{tab:results_interactive_visual}
\vspace{-0.7em}
\end{table}

\begin{table}[t]
    \centering
    \small
    \setlength{\tabcolsep}{4.75pt} 
    \begin{tabular}{cccc}
    \toprule
    $\mathcal{NG}$ & $\mathcal{ET}$ & $\mathcal{AG}$ & \textbf{Acc. \%} \\
    \midrule
    GPT-4o & GPT-4o & GPT-4o& 19.2\\
    GPT-4o mini & GPT-4o & GPT-4o& 0.0 (\textcolor{red}{\textbf{$\downarrow$}19.2}) \\  
    GPT-4o & GPT-4o mini & GPT-4o& 16.5 (\textcolor{red}{\textbf{$\downarrow$}2.7})\\
    GPT-4o & GPT-4o & GPT-4o mini& 17.1 (\textcolor{red}{\textbf{$\downarrow$}2.1}) \\  
    \bottomrule
    \end{tabular}
    \vspace{-0.5em}
    \caption{Performance impact of using different models for $\mathcal{NG}$, $\mathcal{ET}$, and $\mathcal{AG}$ under the Interactive Visual Access setting evaluated on AssistantBench dev split.}
    \label{tab:assistant_bench_analysis}
    \vspace{-1em}
\end{table}

\begin{figure*}
    \centering
    \includegraphics[width=1.0\linewidth]{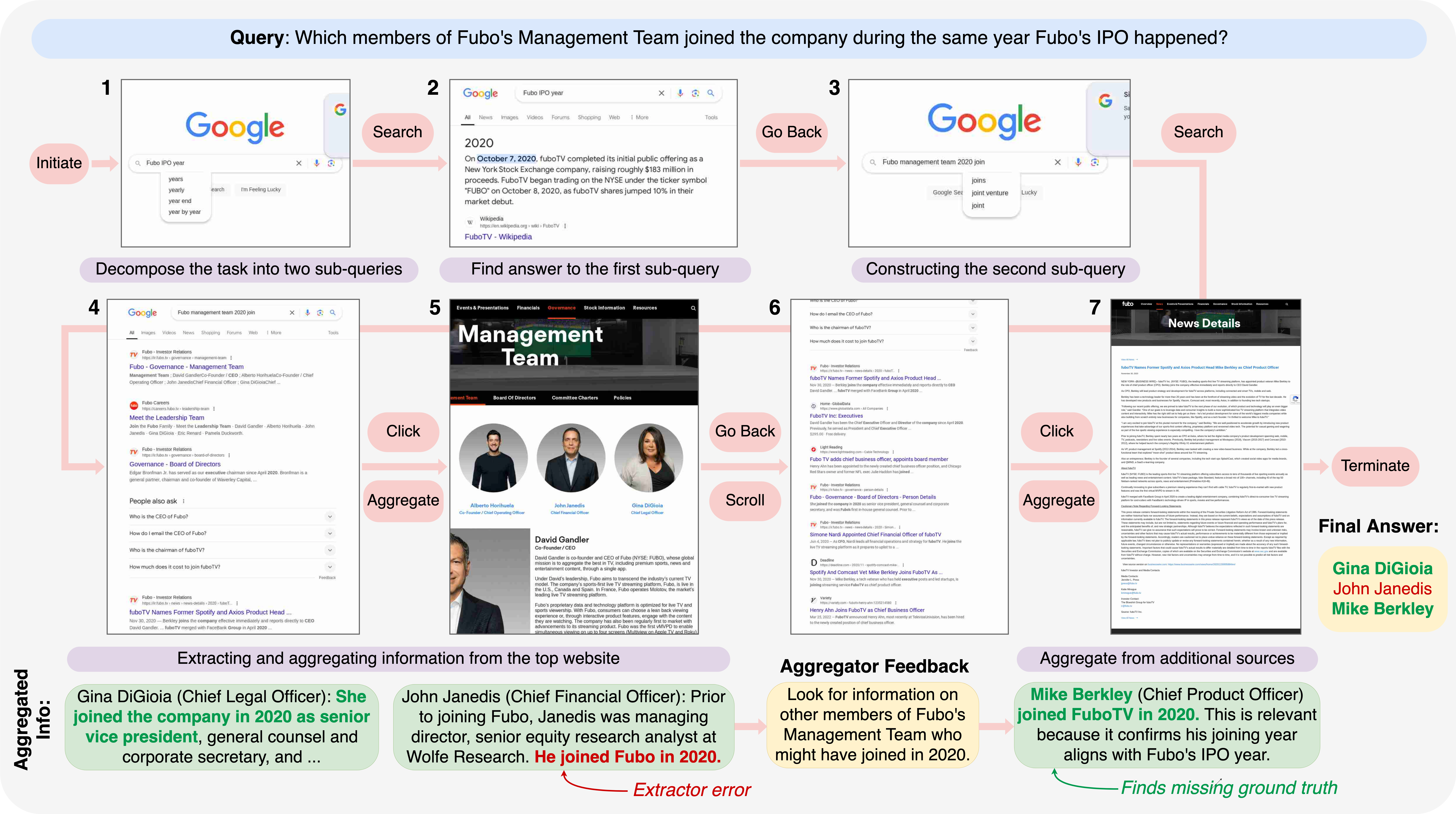}
    \vspace{-1.5em}
    \caption{An illustrative example of \textsc{Infogent} in the \textit{Interactive Visual Access} setting for a query from AssistantBench. In steps 1→4, $\mathcal{AG}$ accurately the identifies the IPO year (2020) and searches for the management team from that year. In step 5, while $\mathcal{ET}$ correctly identifies Gina DiGioia, it incorrectly extrapolates that John Janedia joined in 2020, even though his past affiliations were only mentioned up to that year. However, $\mathcal{AG}$'s feedback to ``look for other members'' improves the answer coverage by discovering Mike Berkley, whose name was not listed on Fubo's current web page, in an external news article (in step 7) noting his appointment as Chief Product Officer in 2020.}
    %First, \textsc{Infogent} queries Google to find Fubo's IPO year, aggregating "2020" (1→2). It then searches for the 2020 management team (2→3), retrieves relevant data after visiting a relevant link(3→4), and, following $\mathcal{AG}$ feedback, seeks more members from 2020 (4→5). Returning to the results (5→6), it selects another link and aggregates additional information (6→7).}
    \label{fig:infogent_qualitative}
    \vspace{-0.8em}
\end{figure*}

\subsection{Analysis}
\label{sec:infogent_analysis}
%Since \textsc{Infogent} directly interacts with websites, human monitoring is necessary to prevent potentially harmful outcomes. We conducted several qualitative experiments, and this section summarizes our findings.
%Here, we present a qualitative analysis of \textsc{Infogent} alongside examples

\subsubsection{Different Models for $\mathcal{NG}$, $\mathcal{ET}$ and $\mathcal{AG}$}
\label{sec:using_diff_models}

We conduct ablation experiments on \textsc{Infogent} under the interactive visual access setting to investigate which component, $\mathcal{NG}$, $\mathcal{ET}$ or $\mathcal{AG}$, is most dependent on the underlying model's capabilities. For this study, we evaluate the performance when using GPT-4o mini instead of GPT-4o for each component separately. Table \ref{tab:assistant_bench_analysis} shows the results on the AssistantBench dev set. We see that the navigator is most reliant on the underlying model, with final accuracy dropping to zero when GPT-4o mini is used for $\mathcal{NG}$. In comparison, using GPT-4o mini for both $\mathcal{ET}$ and $\mathcal{AG}$ results in relatively smaller performance drops of 2.7\% and 2.1\% respectively.

\subsubsection{Distribution of Actions Taken by $\mathcal{NG}$}
%We investigate the frequency of different actions taken by the Navigator $\mathcal{NG}$ in the \textit{Interactive Visual Setting} on the AssistantBench test set. Firstly, we see that 61\% of the instances successfully terminated the navigation, with the remaining leading to timeouts or failures. For these examples, the top-5 actions taken were \texttt{CLICK}, \texttt{AGGREGATE}, \texttt{GO BACK}, \texttt{TYPE} and \texttt{PRESS ENTER} corresponding to 3.40, 3.02, 2.25, 2.01 and 1.32 times per task respectively.
Analyzing the action frequencies of $\mathcal{NG}$ in the Interactive Visual Setting on the AssistantBench test set, we found that 61\% instances successfully terminated navigation, while the remainder resulted in timeouts/failures. The top five actions per task, with their average usage, were \texttt{CLICK} (3.40), \texttt{AGGREGATE} (3.02), \texttt{GO BACK} (2.25), \texttt{TYPE} (2.01), and \texttt{PRESS ENTER} (1.32).

%(These statistics are for data points where the $\mathcal{NG}$ terminated information seeking, i.e. 111/181 examples) \texttt{AGGREGATE} was  triggered 335 times (3.02 times per query) and \texttt{GO BACK} for 249 times (2.25 per query), implying that the framework enabled the model to traverse diverse websites while aggregating information. Further \texttt{CLICK, PRESS ENTER and TYPE} were used 3.40, 1.32 and 2.01 times per query. 

In the Direct API-Driven setting, \texttt{SEARCH} initiates a new query, and \texttt{AGGREGATE} involves website scraping for extraction and aggregation. For FanOutQA, \texttt{SEARCH} and \texttt{AGGREGATE} were used an average of 7.44 and 5.65 times per task, respectively; for FRAMES, these actions averaged 10.4 and 5 times per task, respectively. The higher frequency of \texttt{SEARCH} over \texttt{AGGREGATE} indicates that the navigator more often updates its search query,due to irrelevant results or because the required information is directly available in snippets, rather than extracting information.

%Under the Direct API-Driven setting, among the action available (shown in  Table \ref{tab:direct_api_action}), \texttt{SEARCH} corresponds searching with a new query and \texttt{AGGREGATE} involves scraping a website for extraction. For FanOutQA, we see that \texttt{SEARCH} and \texttt{AGGREGATE} actions were taken 7.44 and 5.65 times per task respectively. Similarly for FRAMES, \texttt{SEARCH} and \texttt{AGGREGATE} actions were taken 10.4 and 5 times per task respectively. The \texttt{SEARCH} being used more than \texttt{AGGREGATE} implies the navigator chooses to update its search query (incase the results are irrelevant or required information is directly provided in the snippets) more often than extracting relevant information from the website via \texttt{AGGREGATE}.

% Given the task, "Find and read three blogs that explain the Transformers architecture in NLP. Compare the differences and summarize the findings.", starting at "google.com", \textsc{Infogent} successfully visits three different blog posts, aggregates information, and generates a response summarizing the findings from the three sources. We note that existing models such as SeeAct are not designed to perform such tasks. 
% \heng{so the information seeking stopping criteria are decided based on conditions like 'three' in the queries?}

\subsubsection{Qualitative Analysis}
\label{sec:qualitative_analysis}

Manual inspection of ten navigation traces in the Interactive Visual Access setting revealed some failure modes in the three components in \textsc{Infogent}. $\mathcal{NG}$ failed to predict correct actions in 6 out of 10 instances, exhibiting issues such as invalid assumptions during Google searches, ignoring aggregator feedback, and repeatedly triggering identical actions (see Appendices \ref{naf_failure} and \ref{geo_naf_failure}). $\mathcal{ET}$  incorrectly judges information in web page screenshots as task-relevant for 3 out of 10 examples, particularly on information-dense pages with distractions (see Figure \ref{fig:infogent_qualitative} for a detailed walkthrough for one such example). $\mathcal{AG}$ often provides open-ended feedback, complicating further navigation; in 3 out of 10 cases, it gave incorrect feedback or omitted relevant information from memory. Conversely, Figure \ref{fig:infogent_qualitative} illustrates how effective aggregator feedback (between steps 5 and 6) can improve answer coverage by appropriately directing the navigator.

\section{Conclusion and Future Work}
\label{sec:next_steps}

% In this work, we explored the capabilities of existing instruction-following LLMs as backbones for Web agents, namely AutoGPT and SeeAct, on Web navigation tasks for breadth-first, information aggregation. We also constructed a benchmark to evaluate the performance of LLM-based Web agents on Web navigation tasks that involve information aggregation, unlike previous works that focused primarily on depth-first, goal-oriented navigation tasks.
% Our analyses regarding the programmatic and non-programmatic approaches highlight the importance of leveraging feedback during the information seeking process not only to determine the termination point, but also to generate improved LLM responses to user instructions that demand information aggregation. We also identify various failure cases in non-programmatic agents where LLMs either fail to circumvent pop-up windows or exhibit an early exit behavior.

%In this work, we assessed the performance of instruction-following LLMs and LMMs, including AutoGPT and SeeAct, as Web agents in breadth-first web navigation tasks focused on information aggregation. We developed a benchmark to evaluate LLM-based Web agents, emphasizing information aggregation, a shift from the typical depth-first, goal-oriented tasks. Our findings underscore the importance of feedback in the information-seeking process to determine the stopping point and enhance LLM responses for tasks requiring information aggregation. Additionally, we propose a new multimodal web agent that can perform information seeking while navigating through the internet with backtracking.

In this work, we introduce \textsc{Infogent}, a novel modular framework for web information aggregation. Through the use of separate Navigator, Extractor and Aggregator components, our approach can incorporate both tool-based LLMs and interactive web agents to handle different information access settings. Experiments demonstrate \textsc{Infogent}'s superior performance over a state-of-the-art multi-agent search framework under Direct API-Driven Access and existing information-seeking web agents under Interactive Visual Access settings. %We demonstrate \textsc{Infogent}'s effectiveness under both Direct API-Driven Access and Interactive Visual Access. 
Future work will incorporate evaluation on a wider variety of web information aggregation tasks. We also plan to explore measuring the diversity and coverage of aggregated information, and to assess ``information sufficiency'' as a criterion for terminating the information-seeking process.

%Further, we plan to explore devising a new metric that explicitly measures the diversity or coverage of the aggregated information, along with judging ``information sufficiency'' as a criteria for termination of the information seeking process.

\section*{Acknowledgement}

We would like to thank members of the BlenderNLP group for valuable feedback and comments. We are grateful to Ori Yoran for helping with making the submission to AssistantBench leaderboard. This research is based upon work supported by DARPA ITM Program No. FA8650-23-C-7316 and DARPA SemaFor Program No. HR001120C0123. The views and conclusions contained herein are those of the authors and should not be interpreted as necessarily representing the official policies, either expressed or implied, of DARPA, or the U.S. Government. The U.S. Government is authorized to reproduce and distribute reprints for governmental purposes notwithstanding any copyright annotation therein.

\section*{Limitations}
\paragraph{Navigation Challenges:} Navigation plays a pivotal role in the success of \textsc{Infogent}. As highlighted in Table \ref{tab:assistant_bench_analysis}, replacing GPT-4o with GPT-4o-mini led to a complete drop in accuracy, emphasizing the need for more effective navigation models. Existing models also struggle with diverse bottlenecks that arise during web navigation, such as solving captchas, indicating room for improvement in their robustness.

\paragraph{Dependency on GPT-4:} While \textsc{Infogent} demonstrates effective collaboration between agents when leveraging high-performing models like GPT-4, the significant performance decline with GPT-4o-mini reveals an over-reliance on GPT-4's capabilities. This underscores the importance of developing open-source models capable of replicating such web navigation proficiency.

\paragraph{Dataset Limitations:} Although \textsc{Infogent} operates as a fully automated framework, the process of information aggregation on the web remains inherently subjective. In our work, we had to rely on multi-hop QA datasets due to the absence of real-world datasets that capture the nuances of subjective information aggregation. Designing appropriate evaluation metrics for such tasks remains a complex challenge, warranting further exploration.

\paragraph{Web's Dynamic Nature:} The constantly evolving nature of the web adds another layer of complexity to information aggregation. Time-sensitive information is prone to changes, and documents are often not updated in a timely manner. Without good SEO practices, outdated content can surface frequently. For large language models (LLMs) to aggregate reliable information, they must account for the relevance and recency of the content they encounter.

\section*{Ethics Statement:}
Automating web navigation introduces several ethical and security challenges. Agents interacting with websites may unintentionally breach terms of service or activate security measures, such as captchas, as previously mentioned. Additionally, there is a risk of accessing or utilizing sensitive or restricted information inadvertently, underscoring the need for stronger ethical guidelines and security protocols within the \textsc{Infogent} framework.

\bibliography{main}
\bibliographystyle{acl_natbib}
\clearpage
\appendix
\section{Appendix}

\subsection{Navigation Failures}
\label{naf_failure}
\begin{figure}[!htb]
    \centering
    \includegraphics[width=\linewidth]{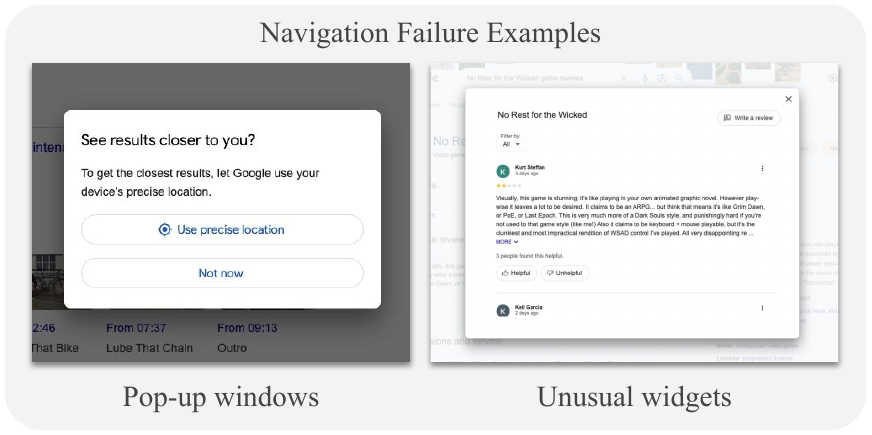}
    \vspace{-15pt}
    \caption{\textsc{Infogent} navigation error examples. The navigator falls in dead loops when encountered unusual web elements, such as pop-up windows asking for sharing locations (left) or ``answer cards'' occasionally appeared at the top of Google search results (right).}
    \label{fig:seeker_nav_err_example}
\end{figure}

The Navigator is a critical component of \textsc{Infogent}. The dynamic nature of the web, especially with its constant updates and varying structures, makes this a particularly challenging task. Navigation failures manifest in multiple forms, including but not limited to pop-ups, AI-generated overviews, captchas, and other interactive elements. While these features are designed to enhance user experience, they also introduce significant barriers for a web agent attempting to navigate efficiently. These obstacles can disrupt the flow of information gathering, making it difficult to access or retrieve data accurately. Samples of web navigation failures are shown in Figure \ref{fig:seeker_nav_err_example}.

\subsection{Geo-Navigational Queries}
\label{geo_naf_failure}

We particularly observed that \textsc{Infogent} struggles with handling geo-navigational queries in AssistantBench. These queries often require precise spatial awareness and the ability to interact with dynamic map interfaces like Google Maps. For example, a query such as ``Which gyms near Tompkins Square Park (within 200m) offer fitness classes before 7am?'' demands not only the retrieval of location-based data but also filtering of relevant details based on distance and time constraints.

In such cases, the model must effectively parse geographic information and interact with Google Maps to identify specific venues within the given parameters. However, this task relies heavily on the Navigator to accurately traverse and manipulate the map interface, which proves to be a significant challenge for current models. Google Maps' dynamic and interactive nature makes it difficult for web agents like \textsc{Infogent} to seamlessly navigate and extract relevant data without human-like intuition. Consequently, handling geo-navigational queries requires sophisticated mechanisms for interpreting spatial data and overcoming the navigational hurdles posed by interactive web platforms. Particularly these queries cause pop-ups like the left one in Figure \ref{fig:seeker_nav_err_example}. 

\begin{table}[ht]
\begin{center}
%\begin{adjustbox}{width=1.0\textwidth}
\begin{tabular}{p{19em}}
\toprule

\tiny{\texttt{\textbf{\hspace{15em}\underline{Task}} \newline
I need your help in evaluating an answer provided by an LLM against a ground truth answer for a given question. Your task is to determine if the ground truth answer is present in the LLM's response. Please analyze the provided data and make a decision.}}\\

\tiny{\texttt{\textbf{\hspace{14em}\underline{Instructions}} \newline
1. Carefully compare the ``Predicted Answer'' with the ``Ground Truth Answer.''\newline
2. Consider the substance of the answers - look for equivalent information or correct answers. Do not focus on exact wording unless the exact wording is crucial to the meaning.\newline
3. Your final decision should be based on whether the meaning and the vital facts of the ``Ground Truth Answer'' are present in the ``Predicted Answer.''}}\\

\tiny{\texttt{\textbf{\hspace{14em}\underline{Input Data}} \newline
- Question: \{question\} \newline
- Predicted Answer: \{predicted\} \newline
- Ground Truth Answer: \{answer\}}}\\

\tiny{\texttt{\textbf{\hspace{14em}\underline{Output Format}} \newline
You should only respond in JSON format as described below and ensure the response can be parsed by Python json.loads.\newline
Response Format: \newline
\{\{ \newline
\hspace{5em}``Explanation'': ``(How you made the decision?)'', \newline
\hspace{5em}``Decision'': ``TRUE'' or ``FALSE'' \newline
\}\}}}\\

\bottomrule
\end{tabular}

%\end{adjustbox}
\caption{Evaluation task for comparing an LLM's predicted answer with a ground truth answer.}
\label{table:evaluation_task}
\end{center}
\end{table}

\begin{table*}[ht]
\begin{center}
%\begin{adjustbox}{width=1.0\textwidth}
\begin{tabular}{p{40em}}
\toprule

\tiny{\texttt{\textbf{\hspace{35em}\underline{Navigator}} \newline \newline
You are an assistant aiding an information aggregation process designed to gather relevant information from the web given a user task. You are provided access to a search tool that you can use to access the web. Your goal is to ensure diversity in the gathered information, so you might want to look at multiple websites in the search results. \newline \newline
You will work in conjunction with an aggregator assistant (which runs as part of the ``extract'' tool) that keeps track of information aggregated and will give feedback to you. It will also let you know how many iterations of calling ``extract'' are left and how many passages it has aggregated so far. You should only visit websites that you think will contain information relevant to user task. If a website does not contain any relevant information, you can skip it. DO NOT visit a website that you have already visited before. \newline \newline
You can leverage the web search multiple times, so that information can be aggregated information over multiple queries. You can decide to stop if aggregator assistant tells you so or if you keep running into a loop. You can simply terminate at the end with a message saying aggregation is done.\newline \newline
Below is the user task. \newline
Task: \{user\_task\}
}}\\

\tiny{\texttt{\textbf{\hspace{35em}\underline{Extractor}} \newline \newline
Website Data: \{data\} \newline \newline
From the above text, extract relevant information for the following task: \{user\_task\}.\newline \newline
You must return the extracted information in the form of a list of paragraphs. Each paragraph should NOT be longer than 8 sentences. Only include the information that is relevant to the provided task. You can extract upto 2 paragraphs ONLY. If the text does not contain any relevant information, you can just return an empty list. 
\newline \newline
You should only respond in JSON format as described below and ensure the response can be parsed by Python json.loads.\newline
Response Format: \newline
\{\{ \newline
\hspace{5em}
``paragraphs'': [list of paragraphs relevant to the task] \newline
\}\}}}\\

\tiny{\texttt{\textbf{\hspace{35em}\underline{Aggregator}} \newline \newline
You are an information aggregation assistant designed to aggregate information relevant to the given user task. Your goal is to ensure diversity in the gathered information while ensuring they are ALL relevant to the user task. Make sure to not gather duplicate information, i.e. do not add redundant information to what you have already aggregated. You can decide to stop aggregating when you decide you have information to address the user task. Also, you can aggregate only \{num\_to\_aggregate\} items in the list and should signal to stop when you have aggregated \{num\_to\_aggregate\} items. \newline \newline
From the above text, extract relevant information for the following task: \{user\_task\}.\newline \newline
You will be provided with a set of passages collected from a website by a navigator assistant. You need to decide whether any of the provided information should be added to the aggregated information list. You have the option to ignore and not add any of the provided passages to the aggregated information list. Also, you should provide feedback to the navigator assistant on how to proceed next. The navigator assistant cannot see the information aggregated, so be clear and specific in your feedback.  You should instruct the navigator to terminate if enough information has been aggregated. You have a maximum of \{num\_iterations\} iterations overall, after which the information aggregated will be automatically returned. \newline \newline
Current Iteration Counter: \{counter\}\newline 
User Task: \{user\_task\}\newline 
Information Aggregated so far: \{aggregated\_list\}\newline 
Provided information: \{provided\_list\} \newline 
\newline
You should only respond in JSON format as described below and ensure the response can be parsed by Python json.loads\newline
Response Format: \newline
\{\{ \newline
``thoughts'': Your step-by-step reasoning for what actions to perform based on the provided information, \newline
``actions'': [list of actions (generated as a string) to perform. Allowed actions are: REPLACE(existing\_id, provided\_id) if passage existing\_id in aggregated information should be replaced by passage provided\_id from provided information and ADD(provided\_id) if passage provided\_id should be added to aggregated information], \newline
``feedback'': Feedback to return to the navigator assistant on how to proceed next. Also, let the navigator assist know how many more iterations are left.\newline
\}\}}}\\
\bottomrule
\end{tabular}

%\end{adjustbox}
\caption{Input prompts for the \textit{Navigator} (top), \textit{Extractor} (middle), and \textit{Aggregator} (bottom) components for the Direct API-Driven Access setting.}
\label{table:programmatic_prompts}
\end{center}
\end{table*}

\begin{table*}[ht]
\begin{center}
%\begin{adjustbox}{width=1.0\textwidth}
\begin{tabular}{p{40em}}
\toprule

\tiny{\texttt{\textbf{\hspace{35em}\underline{Navigator}} \newline \newline
The screenshot below shows the webpage you see. Follow the following guidance to think step by step before outlining the next action step at the current stage: \newline \newline
(Current Webpage Identification) \newline
Firstly, think about what the current webpage is. \newline \newline
(Previous Response and Feedback Analysis) \newline
Secondly, if provided, consider the current response generated for the task along with the feedback. If the response is insufficient, you may need to provide more details to complete the task. For instance, consider revisiting previous search results and exploring other websites to gather additional information. \newline \newline
(Previous Action Analysis) \newline
Then, combined with the screenshot, analyze each step of the previous action history and their intention one by one. Pay more attention to the last step, which may be more related to what you should do next. If the last action involved a TYPE, always evaluate whether it necessitates a confirmation step. \newline \newline
(Screenshot Details Analysis) \newline
Closely examine the screenshot to check the status of every part of the webpage to understand what you can operate with and what has been set or completed. Evaluate the status of every part of the webpage. \newline \newline
(Next Action Based on Webpage and Analysis) \newline
Then, based on your analysis, in conjunction with human web browsing habits and the logic of web design, decide on the following action. Clearly outline which element in the webpage users will operate with as the first next target element, its detailed location, and the corresponding operation. \newline \newline
To be successful, it is important to follow the following rules: \newline
1. You should only issue a valid action given the current observation. \newline
2. If the current webpage has relevant information for the task, trigger AGGREGATE INFORMATION. \newline
3. AGGREGATE INFORMATION is to be used when you think there is factual information that might be useful. \newline
4. You should only issue one action at a time. \newline
5. Press enter after typing a query if needed. \newline
6. Prioritize visiting Wikipedia links over others. \newline
7. Scroll is strictly not an allowed action. \newline
8. Replan if taking the same action repeatedly. \newline
}}\\

\tiny{\texttt{\textbf{\hspace{35em}\underline{Extractor}} \newline \newline
INSTRUCTION: Based on the website's screenshots provided, extract relevant information for the following task: ``{task}".\newline
motivation for aggregating information from this page: ``{search\_motivation}" \newline
Tasks could be multi-hop and information is to be collected over multiple iterations. And the aggregated information from this step will be used for aggregating more detailed information in future steps. \newline
Hence even if the information in the screenshots dont directly answer the query but can help find the answer in future (or has partial information), extract them. \newline
Even if the search motivation has information present, you should extract them from the screenshots. \newline
You should only respond in JSON format as described below and ensure the response can be parsed by Python json.loads. \newline
Response Format: \newline
\{\{ \newline
    ``thoughts": ``details on what the screenshots contain and reason behind the paragraphs aggregated or discarded", \newline
    ``paragraphs": [list of paragraphs extracted from the screenshots relevant to the task. Each paragraph should be detailed (and in string format). For each  entity (name) denote in bracket who they are in context of the task at hand and the motivation for aggregating information (this helps further information aggregation). If there is no relevant information, you can just return an empty list. Don't put your own knowledge into it.], \newline
\}\}}}\\

\tiny{\texttt{\textbf{\hspace{35em}\underline{Aggregator}} \newline \newline
You are an information aggregation assistant designed to aggregate information relevant to the given user task. Your goal is to ensure diversity in the gathered information while ensuring they are ALL relevant to the user task. Make sure to not gather duplicate information, i.e. do not add redundant information to what you have already aggregated. You can decide to stop aggregating when you decide you have enough information to address the user task. Also, you can aggregate only \{num\_to\_aggregate\} items in the list and should signal to stop when you have aggregated \{num\_to\_aggregate\} items. \newline \newline
From the above text, extract relevant information for the following task: \{user\_task\}.\newline \newline
You will be provided with a set of passages collected from a website by a navigator assistant. You need to decide whether any of the provided information should be added to the aggregated information list. You have the option to ignore and not add any of the provided passages to the aggregated information list. Also, you should provide feedback to the navigator assistant on how to proceed next. The navigator assistant cannot see the information aggregated, so be clear and specific in your feedback. You should instruct the navigator to terminate if enough information has been aggregated. You have a maximum of \{num\_iterations\} iterations overall, after which the information aggregated will be automatically returned. \newline \newline
Current Iteration Counter: \{counter\}\newline 
User Task: \{user\_task\}\newline 
Information Aggregated so far: \{aggregated\_list\}\newline 
Provided information: \{provided\_list\} \newline 
\newline
You should only respond in JSON format as described below and ensure the response can be parsed by Python json.loads\newline
Response Format: \newline
\{\{ \newline
``thoughts'': Your step-by-step reasoning for what actions to perform based on the provided information, \newline
``actions'': [list of actions (generated as a string) to perform. Allowed actions are: REPLACE(existing\_id, provided\_id) if passage existing\_id in aggregated information should be replaced by passage provided\_id from provided information and ADD(provided\_id) if passage provided\_id should be added to aggregated information], \newline
``feedback'': Feedback to return to the navigator assistant on how to proceed next. Also, let the navigator assist know how many more iterations are left.\newline
\}\}}}\\
\bottomrule
\end{tabular}

%\end{adjustbox}
\caption{Input prompts for the \textit{Navigator} (top), \textit{Extractor} (middle), and \textit{Aggregator} (bottom) components for the Interactive Visual Access setting.}
\label{table:programmatic_prompts}
\end{center}
\end{table*}

% \begin{figure*}[t]
%     \centering
%     \includegraphics[width=\linewidth]{figures/direct_api_example_v2_detailed.pdf}
%     \vspace{-1em}
%     \caption{A walk-through example of \textsc{Infogent} under the Direct API Access setting. At each step the navigator $\mathcal{NG}$ generates an updated query given the feedback from the aggregator $\mathcal{AG}$, and iteratively performs search and extract procedures according to the updated query at each time step. Within the navigator, we adopt the ReAct \citep{yao2023react} framework's thought-plan-act prompt in order to leverage the LLMs thoughtful planning ability within our \textsc{Infogent}.}
% \label{fig:appendix_infogent_direct_api_detailed}
% \end{figure*}

\end{document}